\documentclass{article}

\usepackage{microtype}
\usepackage{graphicx}
\usepackage{booktabs} %

\usepackage[hyphens]{url}
\usepackage{hyperref}

\usepackage{subcaption}
\usepackage[fleqn]{amsmath}
\usepackage{amsthm}

\usepackage[accepted]{sysml2019}

\sysmltitlerunning{HarDNN: Feature Map Vulnerability Evaluation in Convolutional Neural Networks}

\begin{document}

\twocolumn[
\sysmltitle{HarDNN: Feature Map Vulnerability Evaluation in CNNs}

\begin{sysmlauthorlist}
\sysmlauthor{Abdulrahman Mahmoud}{uiuc}
\sysmlauthor{Siva Kumar Sastry Hari}{nvidia}
\sysmlauthor{Christopher W. Fletcher}{uiuc}
\sysmlauthor{Sarita V. Adve}{uiuc}
\sysmlauthor{Charbel Sakr}{uiuc}
\sysmlauthor{Naresh Shanbhag}{uiuc}
\sysmlauthor{Pavlo Molchanov}{nvidia}
\sysmlauthor{Michael B. Sullivan}{nvidia}
\sysmlauthor{Timothy Tsai}{nvidia}
\sysmlauthor{Stephen W. Keckler}{nvidia}
\end{sysmlauthorlist}

\sysmlaffiliation{uiuc}{University of Illinois at Urbana-Champaign}
\sysmlaffiliation{nvidia}{NVIDIA}
\sysmlkeywords{Reliability, Vulnerability, Errors, SDC, Harden}

\vskip 0.3in %

\begin{abstract}

As Convolutional Neural Networks (CNNs) are increasingly being employed in safety-critical applications, 
it is important that they behave reliably in the face of hardware errors. 
Transient hardware errors may percolate undesirable state during execution, 
resulting in software-manifested errors which can adversely affect high-level 
decision making.
This paper presents \textbf{\textit{HarDNN}}, a software-directed approach 
to identify vulnerable computations during a CNN inference 
and selectively protect them based on their 
propensity towards corrupting the inference output in the presence of a hardware error.
We show that HarDNN can accurately estimate relative vulnerability of a 
feature map (fmap) in CNNs 
using a statistical error injection campaign, and explore heuristics for fast vulnerability assessment.
Based on these results, we analyze the tradeoff 
between error coverage and computational overhead that the system designers can use to employ selective protection.
Results show that the improvement in resilience for the added computation is superlinear with HarDNN. For example, HarDNN improves SqueezeNet's resilience by 10$\times$ 
with just 30\% additional computations.

\end{abstract}

]

\printAffiliationsAndNotice{}  %
\section{Introduction}

CNNs have seen a recent surge in usage across 
many application domains ranging from High Performance Computing (HPC) %
to safety-critical systems 
such as autonomous vehicles and medical devices.
We have also seen a rise in the use of efficient platforms that accelerate 
CNN executions such as GPUs and  
domain-specific accelerators such as the one deployed in Tesla's Full Self-Driving (FSD) System~\cite{Sze2018, drivehw, tesla_ex}.
As CNNs continue to permeate the fabric
of everyday life with increasing utilization in safety-critical applications, it is important that they are resilient to 
transient hardware errors (also known as soft errors).

Studies have shown that hardware errors could have severe
unintended consequences unless the system is designed to detect these 
errors~\cite{OnlineReportToyota1, OnlineReportToyota2}. For example, 
following a series of unintended acceleration events by Toyota vehicles, 
a taskforce following up on a NASA investigation showed that, 
``as little as 
a single bit flip ... could make a car run out of control.'' To mitigate such 
scenarios, hardware in safety-critical systems must fulfill high 
integrity requirements, such as those outlined in the ISO-26262 
standard~\cite{ISO26262}.

While processors deployed in safety-critical systems will employ ECC/parity 
to protect large storage structures (storing weights and intermediate data), 
the level of protection they offer will likely be not sufficient to meet 
the stringent requirements set by standards such as ISO-26262.
Conventional reliability solutions, such as full duplication through
hardware or software, suffer from high overheads in cost, area, power,
and/or performance~\cite{ARMCortexR5, Bartlett2004IBMZ, 
Shye2009PLR}, yet are still commonly used in practice to ensure high resilience. For example, despite the limited power and area constraints of real-time systems, Tesla's FSD system deploys two fully redundant FSD chips 
along with accompanying redundant control logic, power, and peripheral 
packaging on the board for reliability. 
With the goal of developing a reliability solution that is much 
lower cost than full duplication, we seek to understand the underlying 
vulnerability characteristics of CNNs. %
Instead of simply approaching a CNN 
as a single, large computational block, we explore its vulnerability 
at finer granularities (i.e., neurons, feature maps, and layers).
We hypothesize that not all sub-components of a CNN contribute 
equally to the overall network vulnerability, and develop methodologies to 
quantify vulnerability at a finer granularity.
Our results show that 
errors in some feature maps or layers are more likely to corrupt the output 
of a CNN.  Furthermore, we recognize that feature maps are robust to translation effects in 
the input, maintaining higher-level information required
by the CNN for inference, while 
a technique that operates at neuron-level will not have this benefit. 
Based on this advantage and the fact that we can compose the vulnerability
estimates at layer or network level using fmap-level analysis, we focus 
on feature map-level granularity in this work.

One technique commonly used to quantify an application's vulnerability 
to transient errors is error injection experiments. An exhaustive 
study, where one error simulation is performed for a possible hardware 
error in an application state to quantify the effect on the output, is 
often intractable.  
Instead, resilience analyses typically employ a statistical technique 
to limit the number of error simulation runs, while preserving the quality of 
results. We leverage a similar approach to quantify 
the vulnerability of a CNN's component when subjected to 
various transient error models by analyzing the likelihood of a Top-1
misclassification for (classification) models.

In an error injection run, the output of a CNN can be corrupted but the classification might still be correct. To capture the severity of 
output corruption in each injection run, we propose using an alternate
metric that uses the average change 
in cross entropy loss (\textit{$\Delta$loss}). We find that capturing the fine-grain severity metric (instead of binary classification result) 
can produce vulnerability estimates that are comparable whether the classification changes, 
but several times faster (e.g., 10$\times$ for ResNet50).
For an additionally faster method, we explore six heuristics that do not 
perform error injections to estimate vulnerability. We leverage 
activation values and gradient information during inference and 
back propagation. These heuristics provide an additional
tradeoff between accuracy and speed for vulnerability assessment.
We also study the tradeoff between resilience improvement and
increase in computation for added resilience offered by protecting highly
vulnerable feature maps. Results show that a fraction of feature maps 
typically account for a disproportionately large percentage of output corruptions 
(on average, 30\% of feature maps account for 76\% of output corruptions for the studied CNNs). Since each feature map is a convolution
of the input based on an given filter, a low-cost mitigation technique can be 
selective filter (or feature map) duplication.

In summary, this paper presents \textbf{HarDNN}, a highly tunable technique  
to identify and harden the most vulnerable components of a CNN for 
hardware-error-resilient inferences. The following are the main contributions.

\begin{itemize}
    \item We compare various granularities for protection to avoid full 
    CNN duplication. We identify that \textit{feature maps (fmaps)} 
    provide a "sweet spot" for their robustness to translation 
    effects of inputs, and their composability for high-level 
    (e.g., layer-level) protection.
    \item We study the sensitivity of error models on the contribution of a fmap towards the total vulnerability, which we call relative vulnerability. Results show that the relative vulnerabilities of fmaps do not change much with the studied error models.
    \item We introduce $\Delta$loss as an accurate metric to 
    measure vulnerability. $\Delta$loss captures fine-grain perturbations 
    in inference output and converges to the 
    relative vulnerability estimate per fmap with far fewer injections compared
    to the baseline classification-based criterion.
    \item We evaluate multiple non-injection based heuristics for 
    vulnerability estimation, and compare their accuracy and speed to $\Delta$loss.
    \item We study the tradeoff between resilience improvement and
    increase in computation for the resilience offered by protecting highly
    vulnerable fmaps.
    Results show that HarDNN improves resilience of 
    SqueezeNet, for example, by 10$\times$ with just 30\% additional computations.
    
\end{itemize}

\section{Background}
\label{sec:background}

\vspace{-0.1in}
\subsection{Resiliency}
\label{sec:back:DNN}

\begin{figure*}[t]
        \centering
        \includegraphics[width=1.8\columnwidth]{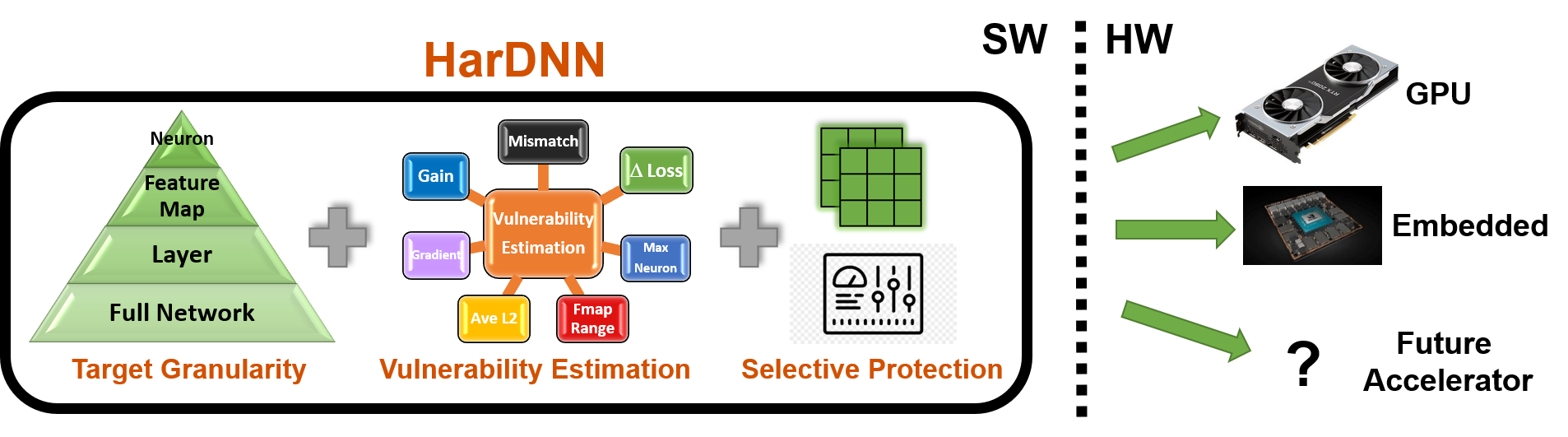}
        \vspace{-0.2in}
        \caption{HarDNN overview. Given a pre-trained model, HarDNN (1) targets a specific granularity for resilience analysis, (2) estimates the relative vulnerability of the components at the chosen granularity, and (3) enables selective protection of the most vulnerable components before model deployment for optimized (real-time) inference.}
        \label{fig:HarDNN_overview}
\end{figure*}

There are two main approaches for resiliency analysis:
\textit{statistical error injection} and \textit{analytical error propagation 
modeling}. Error injection emulates a hardware error by perturbing internal 
program state, and then executing the program to completion to evaluate the 
effect of the error~\cite{lu2015llfi, hari2017sassifi, chang2018hamartia, ApproxilyzerMicro16, MahmoudVenkatagiri2019, FangLu2016, 
FAIL2015, 
Merlin2017,
Relyzer2012,
LiAdve2007b, Sridharan2009}.
Because a program can consist of billions of operations and there 
are a plurality of errors possible for each operation, 
an error injection 
campaign can take a large amount of time and resources to completely 
characterize the resilience of an application. 

Analytical error models attempt 
to reduce the resource intensity of error injection by estimating the 
vulnerability of different operations through higher-level models that take 
into account architecture or domain knowledge~\cite{feng2010shoestring, 
laguna2016ipas, li2018modeling}. In this paper, we investigate both of these 
resiliency analysis approaches for CNNs through injection-based and non-injection-based 
feature map vulnerability assessment schemes.

\subsection{CNN Background}
\label{sec:back:DCNN}
The most fundamental computational unit in a CNN is a neuron 
(or \textit{activation value}).
A \textit{neuron} is the result of a dot product between a filter and an equal sized portion of the input. 
An \textit{output feature map} (or fmap for short) is a plane of many neurons, and is obtained by convolving a filter over an 
input fmap. Mathematically, a convolution is comprised of many dot products, where each dot product 
is composed of many multiply-and-accumulate (MAC) operations. 

A CNN is hierarchically composed of many \textit{convolutional layers},
which are themselves formed of many filters. During an inference, filters are convolved with \textit{input fmaps}
to produce \textit{output fmaps}, where the number of output fmaps correspond 1-to-1
with the number of filters in the layer.
A CNN consists of a series of convolutional 
layers followed by some fully-connected layers. The final layer in the network is 
typically a \textit{softmax} layer, which provides a probability distribution for each 
possible classification the network is trained to predict. The class with the highest 
probability (the \textit{Top-1} class) is the chosen prediction of the network during an inference. 

\section{HarDNN: Design Overview}
\label{sec:approach}

HarDNN is a software-directed resiliency analysis %
technique which identifies and selectively hardens vulnerable computations
of CNN inferences. HarDNN takes as input a pre-trained model, 
estimates the relative vulnerability at a target granularity, and 
hardens the network before deployment using a chosen selective
protection method (e.g., low-level duplication) by protecting the 
most vulnerable components.
Effectively, HarDNN transforms the original CNN model into a 
transient hardware-error-resilient model without any loss of 
classification accuracy. Figure~\ref{fig:HarDNN_overview}
depicts the high-level overview of HarDNN.

In order to effectively analyze and protect a CNN from errors, there are three
fundamental questions that need to be addressed: 
(1) At \textit{what} granularity (i.e., neurons, feature maps, layers) 
should the hardening focus on? 
(2) \textit{Which} subset of the target granularity 
need duplication? 
(3) \textit{How} should the selective protection be implemented?
The rest of this section addresses these questions.

\subsection{Target Granularity}
\label{sec:approach:attributes}
As described in Section~\ref{sec:back:DCNN}, a CNN is composed in a hierarchical manner 
with neurons, feature maps, and layers building up to a full network.
While full network duplication provides high resilience, 
it can incur 2$\times$ runtime or power overheads and 
lowers the throughput offered by the system by half. Such high overheads 
are often prohibitive for safety-critical systems that demand high compute throughput 
and resilience. 
Moreover, full duplication might provide unnecessary over-protection,  
as it is does not consider the contribution of different subcomponents on the 
total system reliability. 
Targeting finer granularities for duplication allows for
effectively allocating resources for reliability, rather than indiscriminate
redundancy in time or space. However, it is critical that the target granularity
also effectively measures vulnerability.
For example, although neuron-level resiliency analysis may provide the most 
fine-grained control for selective hardening, this level of granularity suffers from
a fundamental issue that the neurons are not immune to translational effects of
the input image. Changes to the image orientation or scaled images
can affect the vulnerability contribution of a neuron.

Fmaps, on the other hand, do not suffer from this issue. As long as the CNN is
trained to correctly infer images with such variations in the input (as is typical
in training highly accurate networks), the same fmap is expected to be important 
for similar images. Hence, HarDNN focuses on quantifying vulnerability and hardening 
at the fmap level. 
A benefit of targeting the fmap-level is that the results can be composed to 
perform layer- and network-level vulnerability analysis.

\subsection{Vulnerability Estimation}
\label{sec:approach:heuristics}

Vulnerability of a fmap is defined as the probability of the model's 
output corruption given a transient hardware error during an inference. We refer 
to this quantity as $V_{fmap}$. We can estimate $V_{fmap}$ in two steps using Equation~\ref{equ1}. 
The first step estimates the probability of an error manifestation at the fmap level, 
given a transient hardware error and the second step estimates the probability of 
error propagation of the fmap corruption to the output of the CNN. We refer to these
two quantities as \textit{origination probability} or $OrigP$ and 
\textit{propagation probability} or $PropP$, respectively.
\begin{equation}
V_{fmap} = OriginP \times PropP
\label{equ1}
\end{equation}

$OrigP$ depends on the implementation of the convolution and architecture on 
which it is being run. Assuming that the major storage structures are protected 
in the target hardware platform, most of the errors originate from the 
unprotected computations. Given that MAC operations are 
used to perform a convolution and produce an fmap, we assume that the origination 
probability is directly proportional to the number of MACs in a convolution, without 
loss in generality. In this work, we compute $OrigP$ as a fraction of the number 
of MACs in a convolution to the number of MACs in the entire CNN. Refining this 
quantity based on the hardware platform and implementation optimizations is part 
of our future work.

$PropP$ depends on how the low-level hardware error manifests at the fmap-level 
and how this manifestation propagates through the network to the output. Considering 
a computation-based error model, we assume that an error in a MAC will corrupt 
a single neuron's output in a fmap. We estimate the probability of 
single neuron corruptions propagating to the CNN's output using two classes of techniques -- injection- and non-injection-based.
The first class aims to obtain high accuracy vulnerability estimates. 
The second class aims to estimate the vulnerability fast with zero error injections. 
We describe the techniques in detail in Section~\ref{sec:implementation}. 
We also study the sensitivity of using different neuron-level error 
manifestation models on $PropP$, as described in Section~\ref{sec:methodology:error_model}.

Once we obtain $V_{fmap}$ using Equation~\ref{equ1}, we can estimate the total vulnerability of a CNN model, $V_{CNN}$, using Equation~\ref{equ2}, where $N$ is the total number of fmaps in a CNN.
\begin{equation}
V_{CNN} = \sum_i^N{V_{fmap_i}}
\label{equ2}
\end{equation}

HarDNN aims to estimate the \textit{relative vulnerability} of each of the fmaps in a CNN
(i.e., the contribution of a fmap towards the total CNN vulnerability) to 
address \textit{which} fmaps are the most vulnerable and require protection. The quotients of Equations \ref{equ1} and \ref{equ2} can be used to measure this quantity for an fmap, as shown in Equation~\ref{equ3}.
\begin{equation}
RelV_{fmap_i} = V_{fmap_i} / V_{CNN} 
\label{equ3}
\end{equation}

\subsection{Selective Protection}
\label{sec:implementation:fmap_dup}
Once the relative vulnerabilities of feature maps are gathered, 
HarDNN employs selective duplication to harden the computations 
of the most vulnerable feature maps. 
HarDNN addresses the \textit{how} of selective protection 
by assuming that the filters which correspond to the highly vulnerable 
fmaps can be duplicated. Filter duplication results in two copies of the same logical 
fmap, where any mismatches between the two copies are used to detect errors
during inference and trigger a higher-level system response.
The duplicated fmaps need to be dropped before the inference proceeds
to the next layer. The comparison of the two duplicate feature maps can be performed lazily 
to remove it from the critical path, allowing subsequent layer execution 
to proceed before the output is verified.
HarDNN's highly tunable software-directed
selective protection approach allows the designer to control
the resiliency versus overhead trade-off based on 
the varying resiliency requirements of the system.

\section{HarDNN Vulnerability Estimation Techniques}
\label{sec:implementation}

Accurately measuring PropP exhaustively
would require observing every possible error in every MAC unit of an fmap,
and aggregating the observed error propogation outcome for each error per famp.
This is infeasible, due to the intractably large number of possible error sites.
Instead, we use statistical error injections to accurately obtain
the vulnerability of fmaps (Section~\ref{sec:implementation:injections}). Although tractable,
this may still be slow since statistical significance might require a large number 
of error injections. Section~\ref{sec:implementation:loss} introduces $\Delta$loss
as an injection-based metric which can relatively quickly converges to estimate vulnerability.
We then study how non-injection based heuristics can approximate injection-based vulnerability
in terms of accuracy, but with much less runtime (Section~\ref{sec:implementation:non_injections}). 
Table~\ref{table:heuristics} summarizes all
HarDNN techniques explored for estimating fmap vulnerability.

\subsection{Injection-Based Vulnerability Quantification}
\label{sec:implementation:injections}

The first two metrics used for assessing vulnerability of an fmap are based on injecting an error 
into a CNN during inference, and comparing the outcome of the injection execution
with the error-free execution outcome. The first metric is a binary 
observation of whether the injection resulted in an output
misclassification (compared to the golden, error free inference); this is referred to as a \textit{mismatch}. 
The second metric uses the average delta cross-entropy loss (\textit{$\Delta$loss} for shorthand) 
to measure the vulnerability of an error injection.

\subsubsection{Mismatches}
\label{sec:implementation:mismatches}
We quantify PropP of a specific fmap as 
the fraction of injections that result in a 
classification \textit{mismatch} over 
all error injections performed on the fmap.
An injection run is categorized as a \textit{mismatch} if 
the network misclassifies the input image, as 
compared to the reference label of the image provided by the 
dataset. This is analogous to the Top-1 accuracy metric typically
used in machine learning to assess a neural network's accuracy. 
Error injections that do not alter the Top-1 classification are 
considered \textit{masked}, as their manifestation does not 
change the expected classification of the network.

We deem the mismatch metric to be the most accurate from a resiliency 
standpoint
for computing a fmaps's vulnerability, since the Top-1
category is typically used to interpret the classification
of an input image in an application.
Thus, mismatch forms our ``oracle" metric, which we use
to assess the accuracy of all other metrics' relative fmap
vulnerability.

\subsubsection{$\Delta$Loss: Average Delta Cross Entropy Loss}
\label{sec:implementation:loss}
Although measuring relative vulnerability using mismatches is the gold standard 
for image classification networks, one primary drawback is that mismatches
are relatively rare. Thus, it may take too many injections to obtain fine-grained
differences between fmap vulnerabilities.
A primary insight in this work is to replace the binary view of error propagation 
(represented by mismatch) with a continuous view. To that end, we motivate 
using the average delta cross entropy loss for vulnerability estimation.

Cross entropy loss is typically used to train DNNs during backpropogation 
to improve the prediction accuracy of the network. More generally, it is
used in information theory to measure the entropy between two distributions,
the true distribution and the estimated distribution, by penalizing low confidence
in predictions as well as wrong predictions. Adapting cross-entropy loss to reliability,
we calculate the absolute difference between the cross entropy loss observed
during an error-free inference, and the cross entropy loss observed during an error-injected
inference. This can be expressed as:
\begin{equation}
\Delta \mathcal{L}_{fmap} =  \frac{  \sum_i^N{\mid(\mathcal{L}_{golden} - \mathcal{L}_{i})} \mid } {N} 
\end{equation}
where $\mathcal{L}$ is the cross-entropy loss for an inference, ($\mathcal{L}_{i}$ 
is an error-injected inference, and $\mathcal{L}_{golden}$ represents the golden 
loss for an error-free inference) across $N$ 
total error injections. We use the absolute difference of the loss values to capture 
the \textit{magnitude} of the \textit{relative} loss observed due to an error injection 
as a measure of the vulnerability. The larger the $\Delta$loss, the more vulnerable 
the fmap is estimated to be.

\begin{table}[t]
    \caption{Vulnerability Estimation Techniques}
    \scriptsize
     	\centering
        \renewcommand{\arraystretch}{1.0}
            \begin{tabular}{|c|c|c|}
                    \hline
                      & \textbf{Injection}     &  \\
                      \textbf{Name} & \textbf{Based?}     & \textbf{Brief Description} \\
                    \hline\hline
                    Mismatch        & Yes   & Top-1 Misclassification due to error in fmap \\
                    \hline
                    $\Delta$Loss    & Yes   & Average delta cross entropy loss of fmap \\
                    \hline
                    MaxNeuron       & No    & Max neuron value observe for fmap \\
                    \hline
                    FmapRange       & No    & Range of neuron values observed for fmap \\
                    \hline
                    L2              & No    & Average L2-norm value of fmap \\
                    \hline
                    Gradient        & No    & Average magnitude of gradients for fmap \\
                    \hline
                    Gain            & No    & Analytical model of Top-1 class change \\
                                    &       & for variation in fmap ~\cite{SakrShanbhag2018} \\
                    \hline
                    Mod Gain        & No    & Alternative formulation of Gain analytical model\\
                    \hline
        \end{tabular}

\label{table:heuristics}
\end{table}

%
%
%
%
%
%
%
%
%

%

\subsection{Non-Injection Based Heuristics}
\label{sec:implementation:non_injections}

Non-error injection based heuristics present alternative methods to estimate
vulnerability which can be gathered relatively quickly compared to error injections.
These methods rely on information from a set of error-free inferences to estimate the 
vulnerability of an fmap. 
Our non-injection based heuristics fall under two general categories:
(1) obtaining fmap-level information using observations from the forward pass during an
inference, and (2) performing an additional backward pass (a back-propagation) to
provide additional information via differentiation for vulnerability estimation.
We study a total of 6 non-injection based heuristics.

\textbf{Max Neuron Value}: This simple forward-pass technique assigns an fmap the value of 
the maximal observed neuron value across many sample inputs. Thus, effectively, 
it assumes that errors in feature maps where the activation values
can be high are more likely to affect the outcome. %

\textbf{Feature Map Range}: This technique assigns each fmap the value computed by
finding the difference between the largest and smallest activation value across many 
sample inputs. This ranking scheme takes into consideration that networks
will typically be quantized before 
deployment~\cite{SakrChoi2018,SakrKim2017,SakrShanbhag2018}, constraining their 
dynamic range and, in effect, also reducing the possible observable corruptions in 
the neurons in the feature maps.
Thus, it models the maximal range of error values which may be observed during inference.%

\textbf{Average L2}
The L2-norm calculates
the distance of the vector coordinate from the origin of the vector space. 
Specifically, it is calculated as the square root of the sum of the squared 
vector values. We compute the L2-norm 
of an fmap (vector) averaged across multiple input samples to 
assign this value to the fmap as an estimate of relative vulnerability.

\textbf{Gradient}
One of the key components of 
CNN training is the gradient descent algorithm used to 
update a network's weights. During training, a CNN performs back-propagation to
adjust weights in order to minimize a loss function (a typical loss function is 
cross-entropy loss, discussed above). This is done by obtaining
the gradient value at each weight, and adjusting the weights incrementally
during each training epoch by using the gradient value.
We use a similar technique but adapted to neurons (rather than weights). As neurons are 
differentiable, they too have gradient values which can be used to predict
vulnerability. 
For this technique, we perform a backward pass 
which \textit{only} computes the gradients for each neuron and does not 
modify the network parameters, unlike the backward pass used in the training phase. 
We compute the gradients with respect to the cross-entropy loss at the output.
We use the absolute value of neuron gradients obtained, and average them per fmap across many samples 
from the dataset.

\begin{table*}[ht]
\fontsize{8}{10}\selectfont
\caption{CNNs studied with key topological parameters and training accuracy.}
\scriptsize
     	\centering
        \renewcommand{\arraystretch}{1.0}
            \begin{tabular}{||c|c|c|c|r|c|c|c||}
                    \hline
                      \textbf{Neural} & \textbf{Dataset} & \textbf{Convolutional} & \textbf{Total} & \textbf{Total} & \textbf{Average} & \textbf{Floating Point}           
                      & \textbf{INT8 Quantized} \\
                      \textbf{Network} & \textbf{Name} & \textbf{Layers} & \textbf{Feature Maps} & \textbf{Neurons} & \textbf{Neurons/Fmap} & \textbf{Top-1 Accuracy}           
                      & \textbf{Top-1 Accuracy} \\
                        \hline\hline
                        ResNet50 & ImageNet & 53 & 26,560 & 11,113,984 & 418 & 76.12\% & 75.79\%  \\
                        \hline
                        MobileNet & ImageNet & 52 & 17,056 & 6,678,112 & 391 & 71.87\% & 62.18\%  \\
                        \hline
                        VGG19 & ImageNet & 16 & 5,504 & 14,852,096 & 2,698 & 72.36\% & 72.20\%  \\
                        \hline
                        GoogleNet & ImageNet & 57 & 7,280 & 3,226,160 & 443 & 69.78\% & 69.43\%  \\
                        \hline
                        ShuffleNet & ImageNet & 56 & 8,090 & 1,950,200 & 241 & 69.35\% & 67.01\%  \\
                        \hline
                        SqueezeNet & ImageNet & 26 & 3,944 & 2,589,352 & 656 & 58.18\% & 57.39\% \\
                        \hline
                        AlexNet & ImageNet & 5 & 1,152 & 484,992 & 421 & 56.52\% & 56.04\% \\
                        \hline
                        \hline
                        VGG19 & CIFAR10 & 16 & 5,504 & 303,104 & 55 & 93.34\% & 93.38\% \\
                        \hline
                        AlexNet & CIFAR10 & 5 & 1,152 & 10,752 & 9 & 77.24\% & 77.21\% \\
                        \hline
                        \hline
                        VGG19 & CIFAR100 & 16 & 5,504 & 303,104 & 55 & 71.94\% & 71.89\% \\
                        \hline
                        AlexNet & CIFAR100 & 5 & 1,152 & 10,752 & 9 & 43.87\% & 43.82\% \\
                        \hline
                        
        \end{tabular}

\label{table:dnns}
\end{table*}




\textbf{Gain}
Recent work by Sakr et al.~\cite{SakrKim2017,SakrShanbhag2018} proposed an analytical model
which bounds mismatch probability $p_m$ 
in the context of network quantization. The analysis is based on estimating how much a noise source, at a set of neurons for instance, affects the accuracy of a network. If a set of neurons $\mathcal{A}$ is corrupted element-wise by a noise source of variance $\sigma^2_n$ then: %
\begin{align}
\label{eqn::pm_expectation} p_m\leq \sigma^2_n \mathbf{E}\left[\sum_{i=1,i\neq \hat{y}}^M \frac{\sum_{a\in\mathcal{A}}\left|\partial\left(z_i-z_{\hat{y}}\right)/\partial(a) \right|^2}{\left|z_i-z_{\hat{y}} \right|^2} \right]\end{align}
where $\sigma^2(n_a)$ is the variance of the noise source, $\hat{y}$ is the label predicted by the noise-free network, $\{z_i\}_{i=1}^M$ are that soft outputs, $M$ is the number of classes. 

The set of neurons $\mathcal{A}$ can be defined in a flexible manner and could denote anything from all of the neurons in the network, or the set of neurons in a given layer, and most relevantly, the set of neurons within an fmap. Thus, we define the noise gain of an fmap $F$ as
\begin{align}
    \label{eqn::fm_gain}
    E_{F} = \mathbf{E}\left[\sum_{i=1,i\neq \hat{y}}^M \frac{\sum_{a\in F}\left|\partial\left(z_i-z_{\hat{y}}\right)/\partial(a) \right|^2}{\left|z_i-z_{\hat{y}} \right|^2} \right]
\end{align}
where, from \eqref{eqn::pm_expectation}, we have
\begin{align}
\label{eqn::pm_simple_form}
p_m\leq \sigma^2_n E_F    
\end{align}
 The expectation in \eqref{eqn::fm_gain} can be obtained by taking averages over the training set or even a subset of it with statistically significant size as discussed in~\cite{SakrKim2017}. Furthermore, computing the noise gain in \eqref{eqn::fm_gain} requires derivatives of outputs with respect to neurons, 
 which are readily available thanks to the automatic differentiation packages utilized by deep learning frameworks implementing the back-propagation algorithm. In addition, \eqref{eqn::fm_gain} assumes a pre-trained network with frozen parameters and need only be computed once for all fmaps.

Thus, a natural mechanism for fmap vulnerability estimation
is simply to measure their noise gains. Indeed, if an fmap $F$ has a large noise gain $E_F$, then \eqref{eqn::pm_simple_form} shows that corrupting $F$ with noise will have a large impact on $p_m$. On the other hand, if $E_F$ is small, then corrupting $F$ with noise will have a small impact on $p_m$. In our results, this ranking procedure is referred to as \textit{Gain}. 
%

%

\textbf{Modified Gain}
The analytical results of \textit{Gain} are derived assuming corruption of neurons by independent noise sources. In our work, we also 
consider the case where a neuron is \emph{replaced} by a random scalar belonging to the fmap's dynamic range as discussed 
later in Section \ref{sec:methodology:error_model}. Such setup violates the independence assumption between neuron and noise. 
However, we may still leverage the above theory. Indeed, it can be shown that \eqref{eqn::pm_simple_form} still applies in the 
context of neuron replacement provided the definition of the noise gain is updated as follows:
\begin{align}
\label{eqn::gain_modified_square} E_F = \mathbf{E}\left[\sum_{i=1,i\neq \hat{y}}^M \frac{\sum_{a\in F}a^2\left|\frac{\partial\left(z_i-z_{\hat{y}}\right)}{\partial(a)} \right|^2}{\left|z_i-z_{\hat{y}} \right|^2} \right]\end{align}
This analysis yields a homologous technique to \textit{Gain} which we refer to as \textit{Mod-Gain} in our evaluation.
\section{Evaluation Methodology}
\label{sec:methodology}

We evaluate HarDNN on 11 CNNs across three datasets. Table~\ref{table:dnns}
lists each CNN, along with the number of layers, fmaps, and neurons, and accuracy 
on the respective dataset. We use the PyTorch v1.1 framework~\cite{Pytorch} for
evaluation, and obtained pretrained models for CNNs trained on ImageNet~\cite{ILSVRC15}
from the PyTorch TorchVision repository~\cite{pytorch_imagenet}, and CNNs
trained on CIFAR10/100 from github~\cite{bearpawGitHub}.
All experiments were run on an Amazon EC2 p3.2xlarge 
instance~\cite{AmazonAWS}, which has an Intel Xeon E5-2686 v4 
server processor, 64GB of memory, and an NVIDIA V100 GPU with 
16GB of memory~\cite{NVIDIA2018}. 

\begin{table}
    \caption{Runtime analytical models for vulnerability estimation schemes.}
    
	\fontsize{8}{10}\selectfont
     	\centering
        \renewcommand{\arraystretch}{1.0}
            \begin{tabular}{||c|c||}
                    \hline
                      \textbf{Technique} & \textbf{Runtime Estimate} \\
                    \hline\hline
                    Mismatch & samples * forward * fmaps * inj/fmap \\
                    \hline
                    $\Delta$Loss & samples * forward * fmaps * inj/fmap \\
                    \hline
                    MaxValue & samples * forward \\
                    \hline
                    FmapRange & samples * forward \\
                    \hline
                    Average L2 & samples * forward \\
                    \hline
                    Gradient & samples * (forward + backward) \\
                    \hline
                    Gain & samples * (forward + (backward * (classes - 1))) \\
                    \hline
                    Mod-Gain & samples * (forward + (backward * (classes - 1))) \\
                    \hline
        \end{tabular}

\begin{tabular}{ll}
	\multicolumn{2}{l}{\vspace{-0.15in}} \\
	\multicolumn{2}{l}{Terminology:}  \\
 \hline
 
 samples: & Total number of images used in the ranking set\\
 forward: & Average runtime of a single inference for a CNN \\
 backward: & Average runtime of a single back-propogation for a CNN \\
 fmaps: & Total number of feature maps in a network \\
 inj/fmap: & The number of error injection experiments per feature map \\
 classes: & The total number of output classes 
 
\end{tabular}

\label{table:runtime}
\end{table}

%
\subsection{Error Models}
\label{sec:methodology:error_model}
We model neuron-level manifestations of transient hardware errors that occur during an inference. As described in Section~\ref{sec:approach:heuristics},
we assume that a particle strike to a flip-flop used during a MAC operation during a
convolution will corrupt a single neuron's value. Such low-level errors can manifest 
as single or multiple bit-flips. Additionally, highly optimized inference 
systems typically employ quantization prior to deploying CNNs. 
Such models run significantly faster with 
hardware support for reduced-precision operations, which is prevalent in
GPUs and CPUs. These benefits often come with a small but acceptable loss in 
classification accuracy (reflected in Column 8 of Table~\ref{table:dnns}). 
Based on these consideration, we evaluate $PropP$ using the following three error models.

\textbf{FP-Rand} represents a random (possibly multi-bit) error 
during the computation of a neuron, where the computations are performed
on floating point values. We model this by choosing a random neuron in an
fmap, and substituting the original value with a random value between [-max, max],
where max is the maximum observed neuron value in the fmap across
the training set. FP-Rand limits the effect of an error by bounding it between a range.
Previous work found that inference is highly sensitive to errors in the sign
and exponent bits and a simple output fmap-level range detector can mitigate 
many of the most severe corruptions~\cite{Li2017}.

\textbf{FxP-Rand} considers 8-bit integer (INT8) \textit{fixed-point quantization}, 
which quantizes the CNN based on the range of neuron values observed during training. 
Error injections are performed by choosing a random neuron from an fmap, and substituting the original
value with a randomly selected INT8 value. Thus, FxP-Rand is similar to FP-Rand, but for quantized models.

\textbf{FxP-Flip} models the effect of a single bit-flip 
on a fixed-point quantized neuron, simulating the effect of a particle strike on a flip-flop storing the neuron's output. %

\subsection{Evaluating Vulnerability Estimation Techniques }
\label{sec:eval:mismatch_vs_loss}

For vulnerability analysis, we exclude images that are incorrectly 
classified by the error-free network since our focus is on analyzing the resilience 
of the network during correct execution. After removing incorrectly classified
images from the dataset, we randomize and partition the correctly
classified images into two non-overlapping subsets: an estimation set (ES) 
and a test set (TS), using an 80-20 split. 
We use the TS to perform error injections and generate baseline vulnerability
estimations using the mismatch- and $\Delta$loss-based metrics. 
The ES is used on all metrics (including mismatch and $\Delta$loss), to compare
the fmap vulnerability estimates to the TS.

To quantitatively compare two metrics, we sort the fmaps from most to least vulnerable 
by the respective metric. We compute the cumulative vulnerability of the fmaps in this 
sorted order. This gives us a cumulative vulnerability vs. fmap curve for a given metric, 
from which we can compute Manhattan distance between the two curves for each fmap point.
We use the average Manhattan distance as a measure of the difference between the metrics, where a distance 
close to zero indicates very high correlation.

\subsection{Error Coverage vs.\ Computational Overhead}
\label{sec:eval:coverage_vs_overhead}

For a given set of feature maps, F,  that are duplicated, we define coverage as the cumulative relative vulnerability of those fmaps.  From a developer’s point of view, depending on the metric used, this coverage is the cumulative vulnerability estimate given by that metric on the test set. We refer to this as the \textit{predicted coverage}. The \textit{actual coverage} on the evaluation set, however, could be different. We experimentally measure this coverage by determining the coverage of F on the test set using the loss metric. We validate the predicted coverage against the actual coverage to see how similar they are.

To target a specific coverage value, we use a greedy algorithm. We sort all fmaps from higher to lower vulnerability (based on the metric being considered) and choose the first several fmaps whose relative vulnerability adds up to the targeted coverage. To assess the relative overhead tradeoff, we measure the total number of MAC operations in those selected fmaps as a fraction of the total MAC operations in all fmaps. We use MACs as a reasonable proxy to the actual overhead, while providing some abstraction for the actual hardware used.

\subsection{Heuristic Analysis}
\label{sec:eval:heuristic_analysis}
One of the primary motives of using heuristics is to estimate
vulnerability much faster compared to error injections. We provide analytical models
to compare the different expected runtimes of each vulnerability estimation technique in 
Table~\ref{table:runtime}. The analytical models predict the runtime
based on the the number of samples explored, runtime of a forward/backward pass,
and the number of inj/fmap (for the injection-based techniques). The forward-pass-based
heuristics (MaxValue, FmapRange, AverageL2) are expected to perform the fastest, followed
by the backprop-based heuristics (gradient, gain, mod-gain), and the slowest are
injection-based techniques. In practice, most of these schemes can be parallelized
via batching multiple images together on a single device and 
across multiple devices, which can provide additional speedups 
(not included in the model).

To assess the \textit{accuracy} of the heuristics, i.e., how well they estimate relative vulnerabilities of fmaps, we compare the cumulative vulnerability 
estimates of each heuristic on the ES, and measure the Manhattan distance to an error-injection
estimation provided by the TS. Section~\ref{sec:eval:coverage_vs_overhead} provides additional
detail on quantitatively comparing the accuracy of a heuristic.

\section{Results}
\label{sec:results}

\begin{figure}[t]
        \centering
        \includegraphics[width=.95\columnwidth]{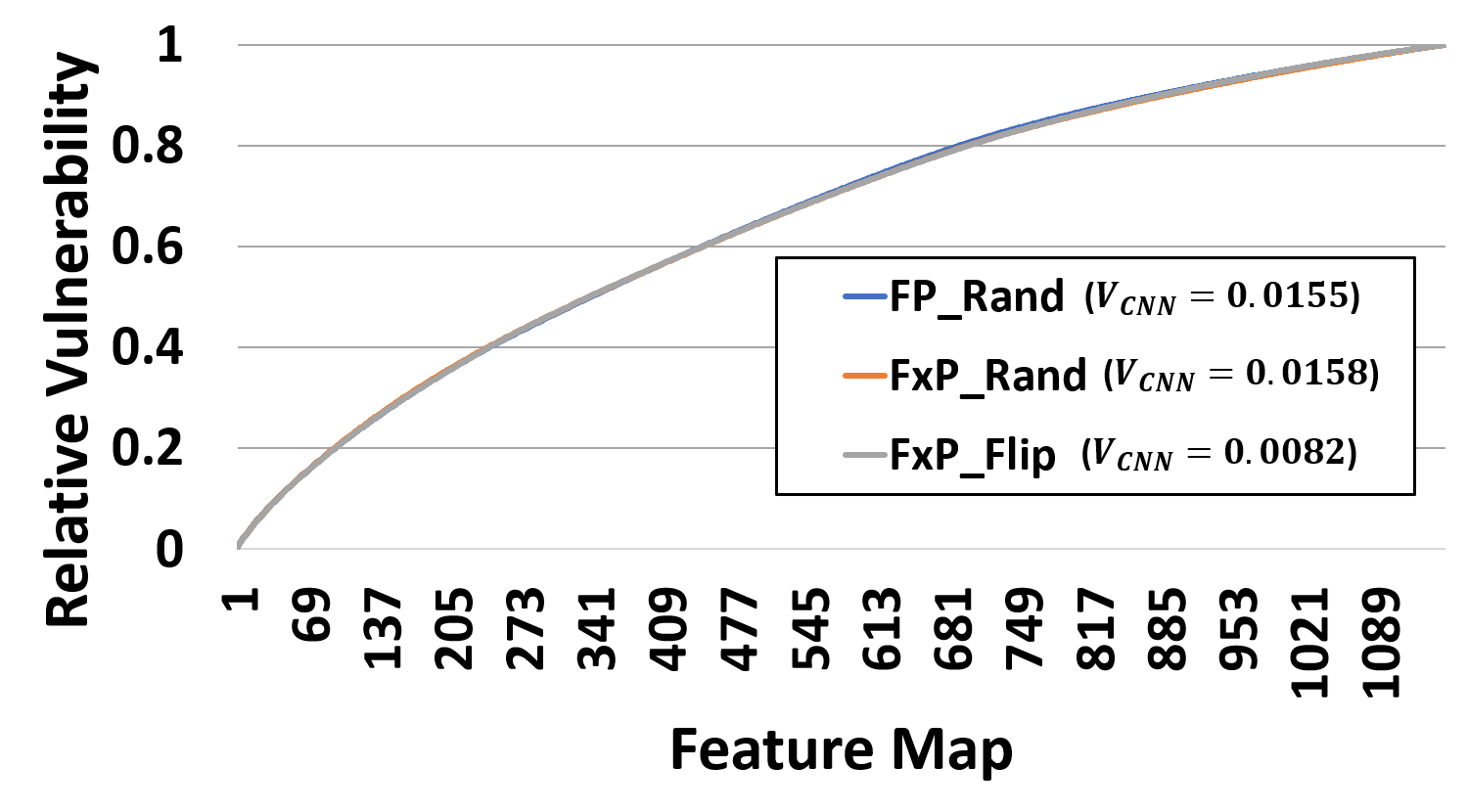}
        \caption{Relative vulnerability of fmaps is similar across error models,
        even with different $V_{CNN}$ (AlexNet-ImageNet)}
        \label{fig:error_model_comp}
\end{figure}

\begin{table}[t]
    \caption{Manhattan Distance between Relative Vulnerability of FP-Rand and FxP-Flip}
    \fontsize{8}{10}\selectfont
     	\centering
        \renewcommand{\arraystretch}{1.0}
            \begin{tabular}{||c|c||}
                    \hline
                    \textbf{Network-Dataset} & \textbf{Distance} \\
                    \hline
                    \hline
                    ResNet50-ImageNet & $0.71\times10^-5$ \\
                    \hline
                    MobileNet-ImageNet & $1.65\times10^-5$ \\
                    \hline
                    VGG19-ImageNet &  $3.95\times10^-5$ \\
                    \hline
                    GoogleNet-ImageNet & $2.18\times10^-5$ \\
                    \hline
                    ShuffleNet-ImageNet & $3.35\times10^-5$ \\
                    \hline
                    SqueezeNet-ImageNet &  $7.76\times10^-5$ \\
                    \hline
                    AlexNet-ImageNet & $2.99\times10^-5$ \\
                    \hline
        \end{tabular}

\label{table:fprand_vs_fxpflip}
\end{table}

\subsection{Error Models Comparison}
\label{sec:results:error_models}
We begin our evaluation by comparing the relative vulnerabilities of fmaps
across different error models. Figure~\ref{fig:error_model_comp} shows
the cumulative relative vulnerability of the fmaps in AlexNet-ImageNet,
where the x-axis is sorted in descending order of relative vulnerability
as measured by the oracle mismatch metric using 12,288 injections per fmap (shorthand: inj/fmap). 
We use the same ordering on the x-axis (based on FxP-Flip vulnerability ordering) for comparison,
and normalize the relative vulnerabilities of fmaps to its respective error model.

Comparing relative vulnerabilities of fmaps show that, regardless of error model,
\textit{an fmap's contribution towards the total network vulnerability is nearly the same},
with the average Manhattan distance between the different errors models for AlexNet 
at $2.4\times10^-5$.
Further, we find that this occurs despite the fact that the absolute total
vulnerability, $V_{CNN}$, differs for different models. FP-Rand and FxP-Rand 
show similar $V_{CNN}$, since both error models have the same dynamic range
and multi-bit perturbation probabilities. FxP-Flip shows lower $V_{CNN}$, attributed
to the less egregious error model of a single-bit perturbation. 
We find that the relative vulnerabilities of fmaps 
are nearly same for different networks for FP-Rand and FxP-Flip.
Table~\ref{table:fprand_vs_fxpflip} shows the distances between the results 
for the two error models for different CNNs are small, despite FxP-Flip displaying 
fewer errors overall compared to FP-Rand.
Given that the relative vulnerability contribution of fmaps is similar across
error models, we focus the remainder analysis of this paper on FxP-Flip.

\begin{figure}[t]
    \centering
    \begin{subfigure}[b] {.45\textwidth}
        \centering
        \includegraphics[width=\textwidth]{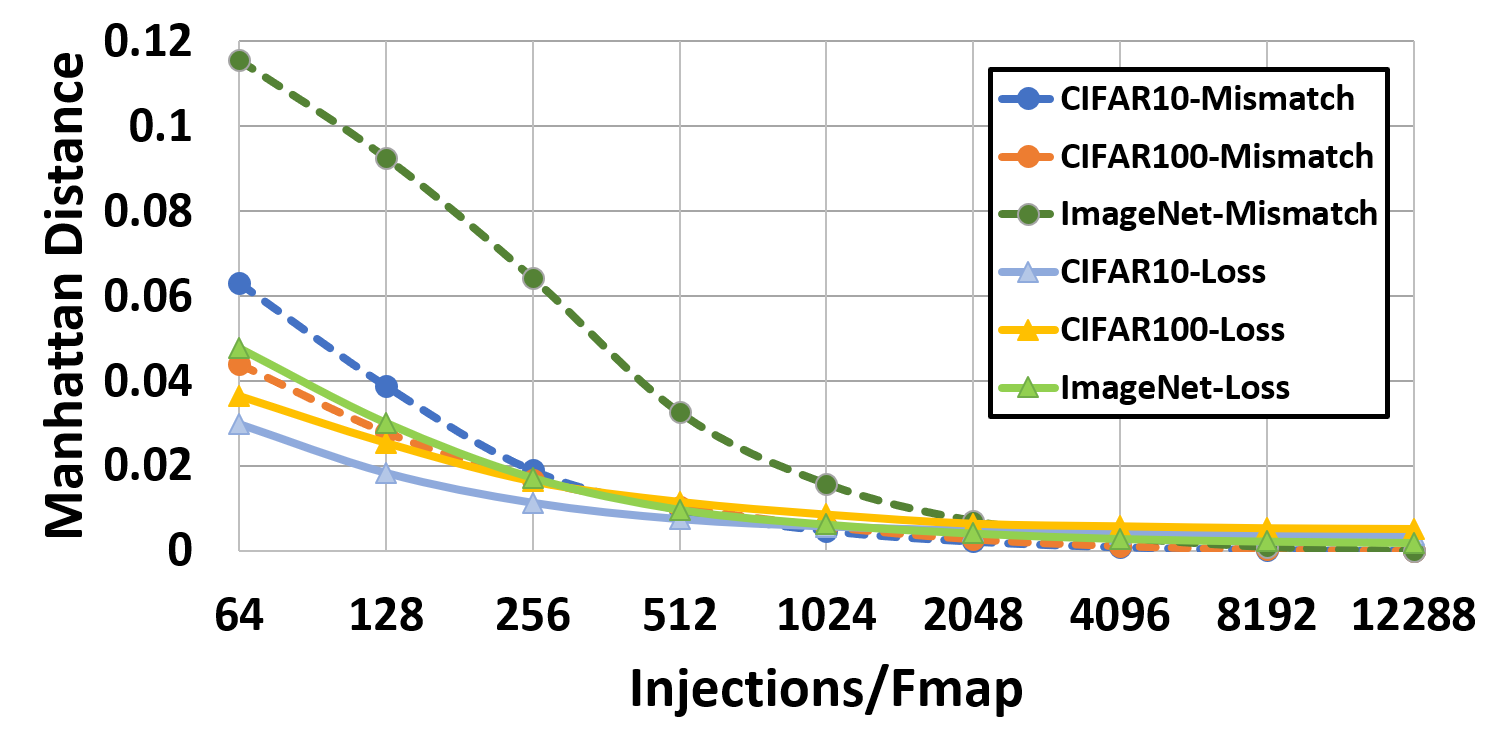}
        \caption{At large inj/fmap, mismatch and loss converge to provide similar fmap vulnerability estimates (AlexNet). Loss converges faster.}
        \label{fig:mismatch_vs_loss:alexnet}
    \end{subfigure}
    \hfill
    \centering
    \begin{subfigure}[b] {.45\textwidth}
        \centering
        \includegraphics[width=\textwidth]{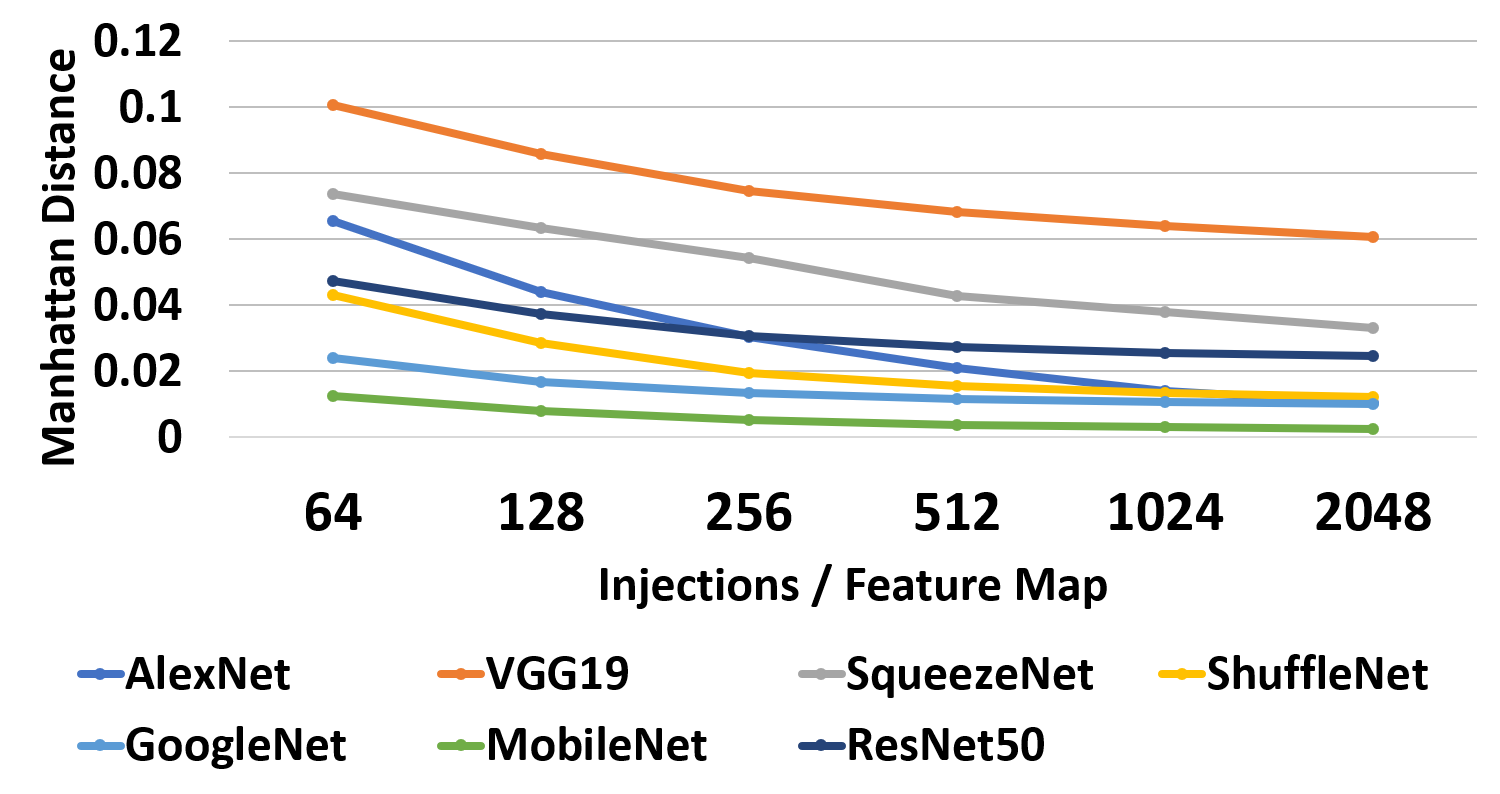}
        \caption{Loss converges to vulnerability estimates with relatively few inj/fmap (across CNNs)}
        \label{fig:mismatch_vs_loss:all_networks}
    \end{subfigure}
    \vspace{-0.1in}
    \caption{Convergence of mismatch and loss
    }
    \label{fig:mismatch_vs_loss}
    \vspace{-.1in}
\end{figure}

\begin{figure}[t]
    \centering
    \begin{subfigure}[b] {.45\textwidth}
        \centering
        \includegraphics[width=\textwidth]{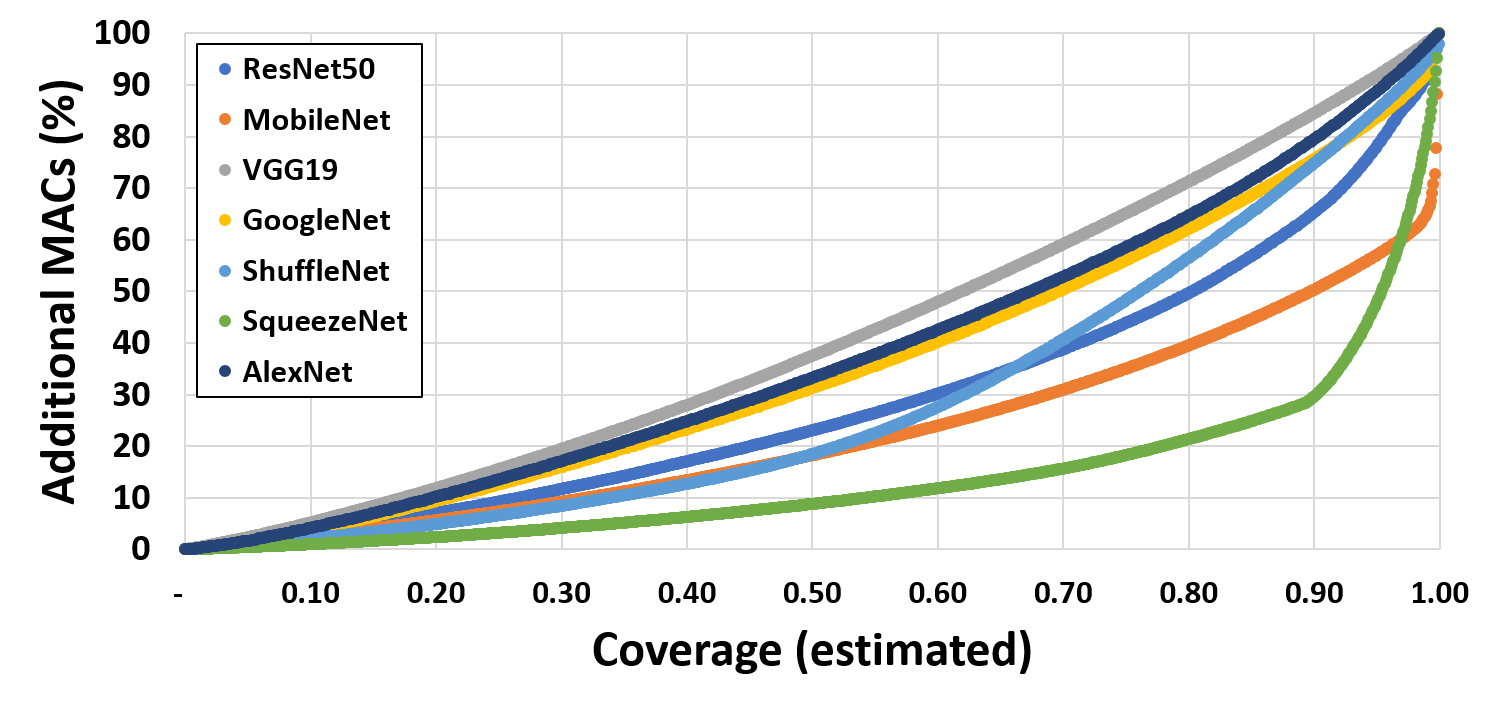}
        \caption{Predicted coverage using $\Delta$Loss vulnerability estimate}
        \label{fig:predicted_coverage}
    \end{subfigure}
    \hfill
    \centering
    \begin{subfigure}[b] {.45\textwidth}
        \centering
        \includegraphics[width=\textwidth]{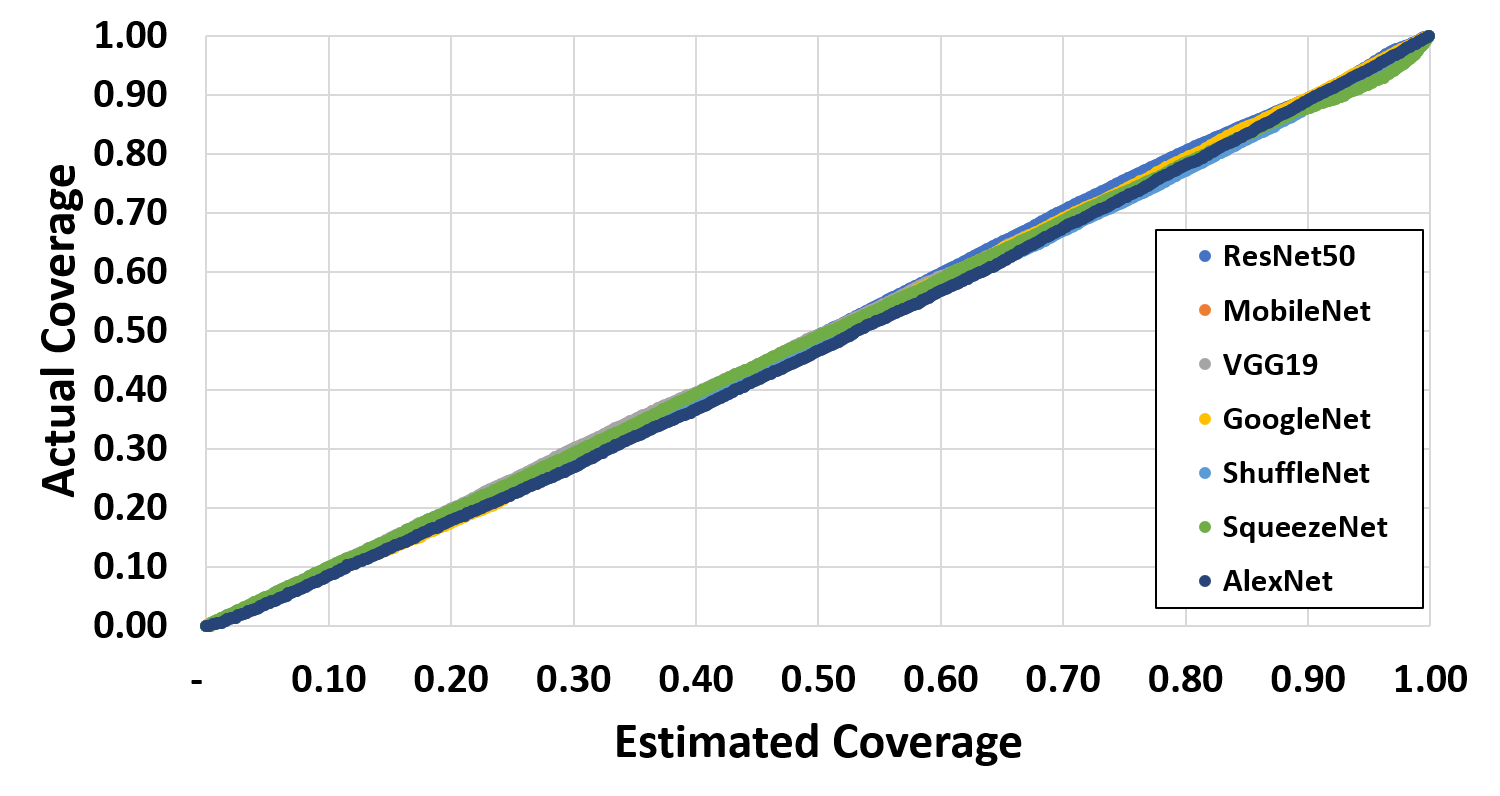}
        \caption{Validation of predicted coverage vs actual coverage}
        \label{results:validation}
    \end{subfigure}
    \vspace{-0.1in}
    \caption{Error coverage vs computational overhead}
    \label{fig:overhead}
    \vspace{-.1in}
\end{figure}

\begin{figure*}[ht]
    \centering
    \begin{subfigure}[b] {.33\textwidth}
        \centering
        \includegraphics[width=\textwidth]{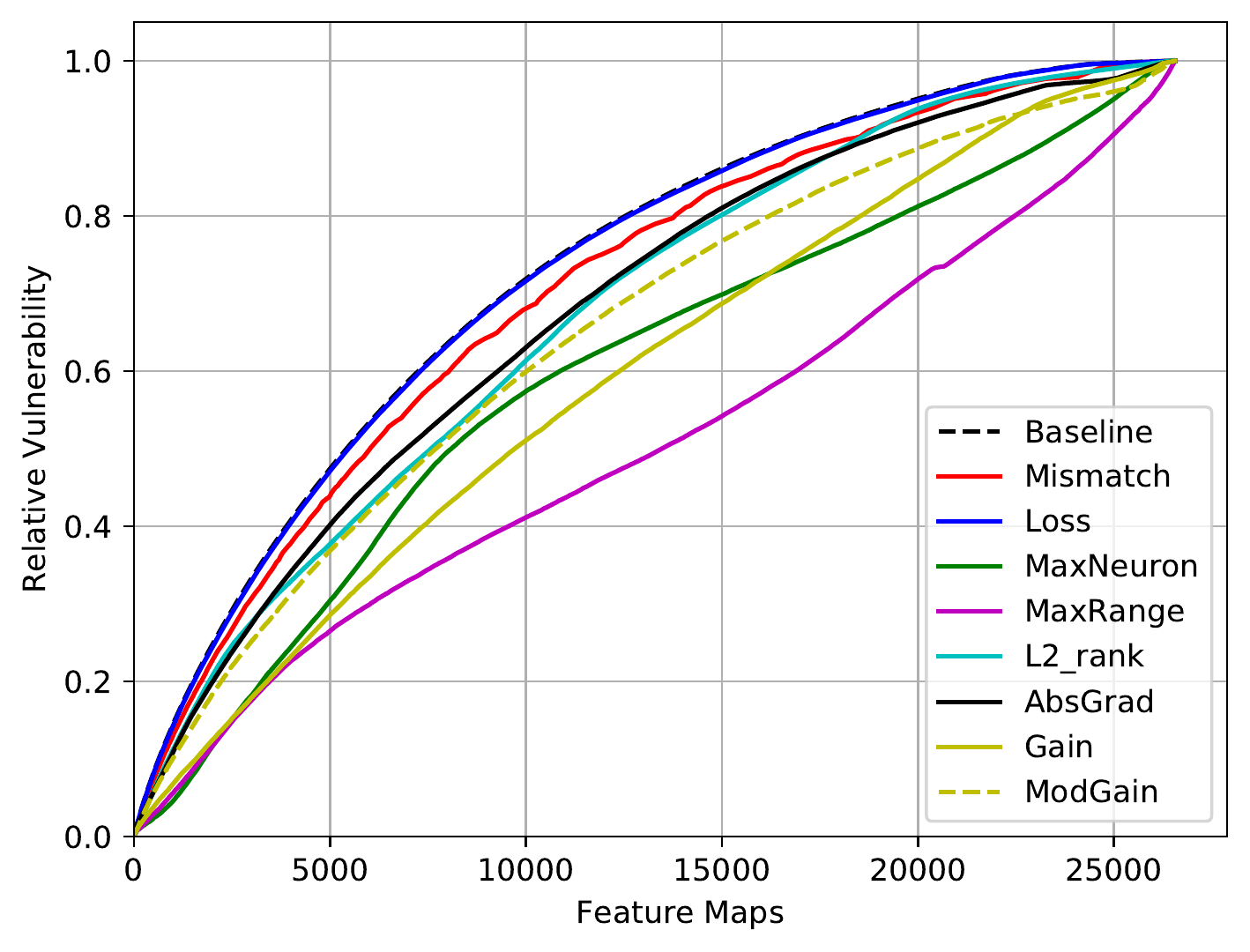}
        \caption{ResNet50-ImageNet}
        \label{fig:accuracy:resnet50}
    \end{subfigure}
    \hfill
    \begin{subfigure}[b] {.33\textwidth}
        \centering
        \includegraphics[width=\textwidth]{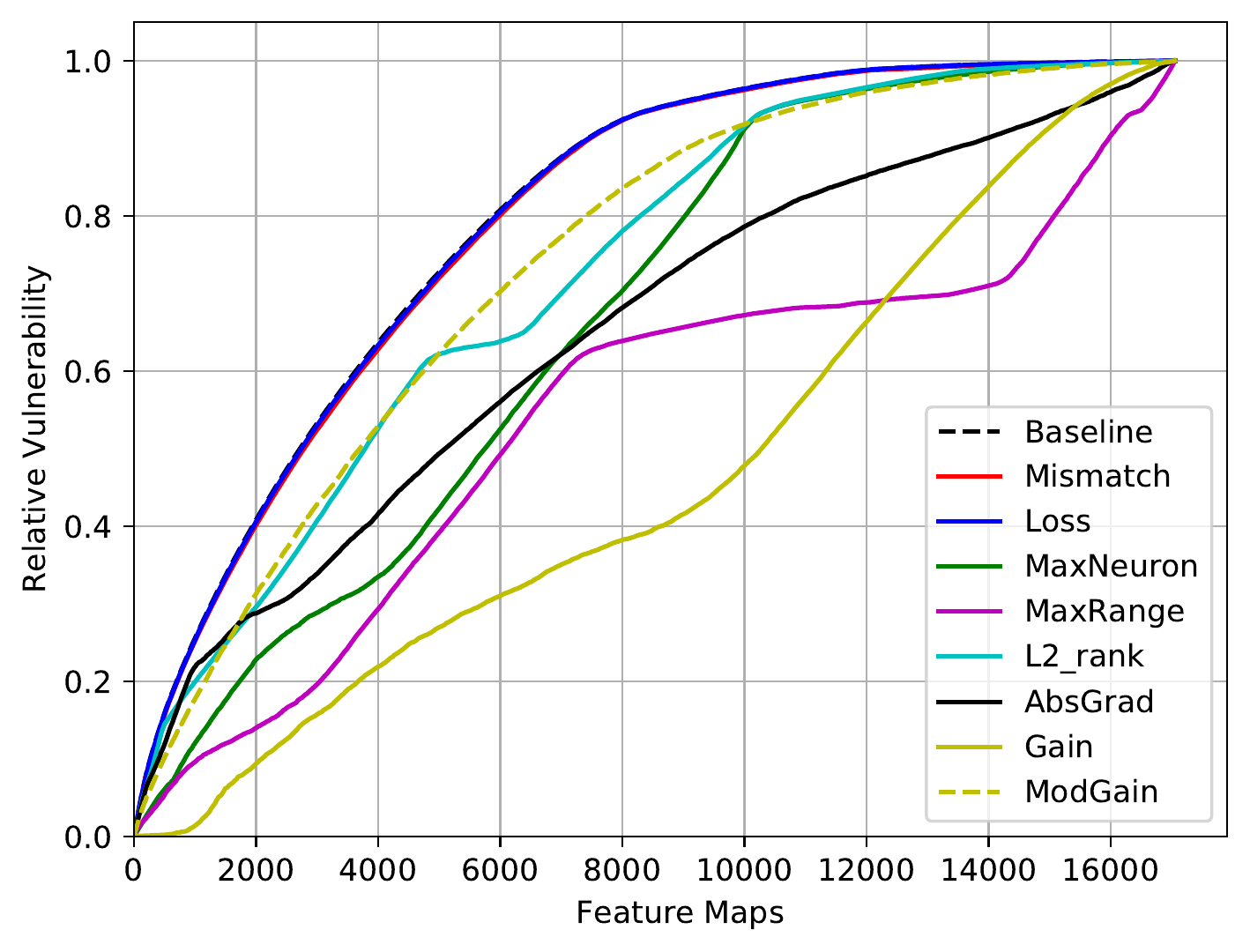}
        \caption{MobileNet-ImageNet}
        \label{fig:accuracy:mobilenet}
    \end{subfigure}
    \hfill
    \begin{subfigure}[b] {.33\textwidth}
        \centering
        \includegraphics[width=\textwidth]{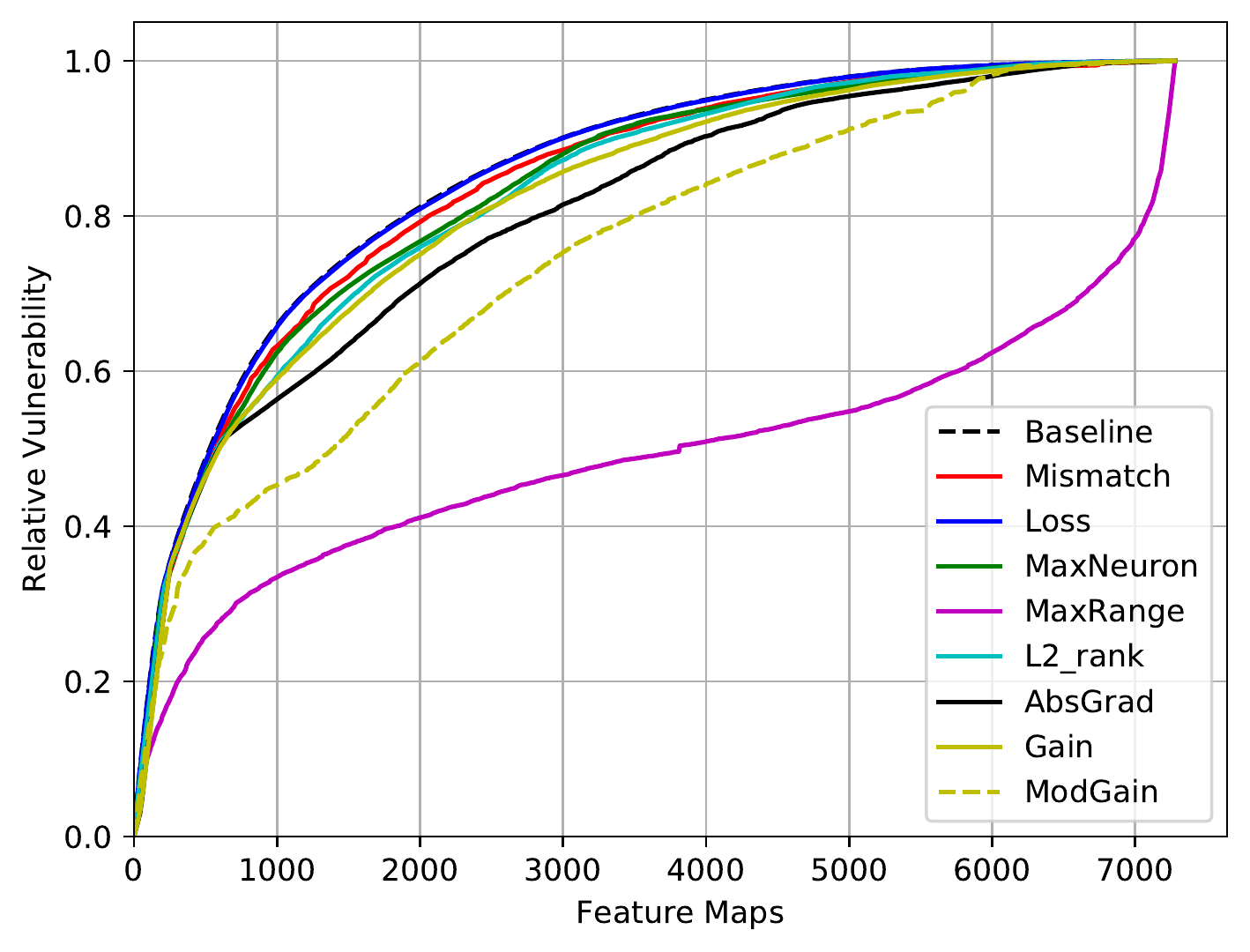}
        \caption{GoogleNet-ImageNet}
        \label{fig:accuracy:googlenet}
    \end{subfigure}
    \centering
    \hfill
    \begin{subfigure}[b] {.33\textwidth}
        \centering
        \includegraphics[width=\textwidth]{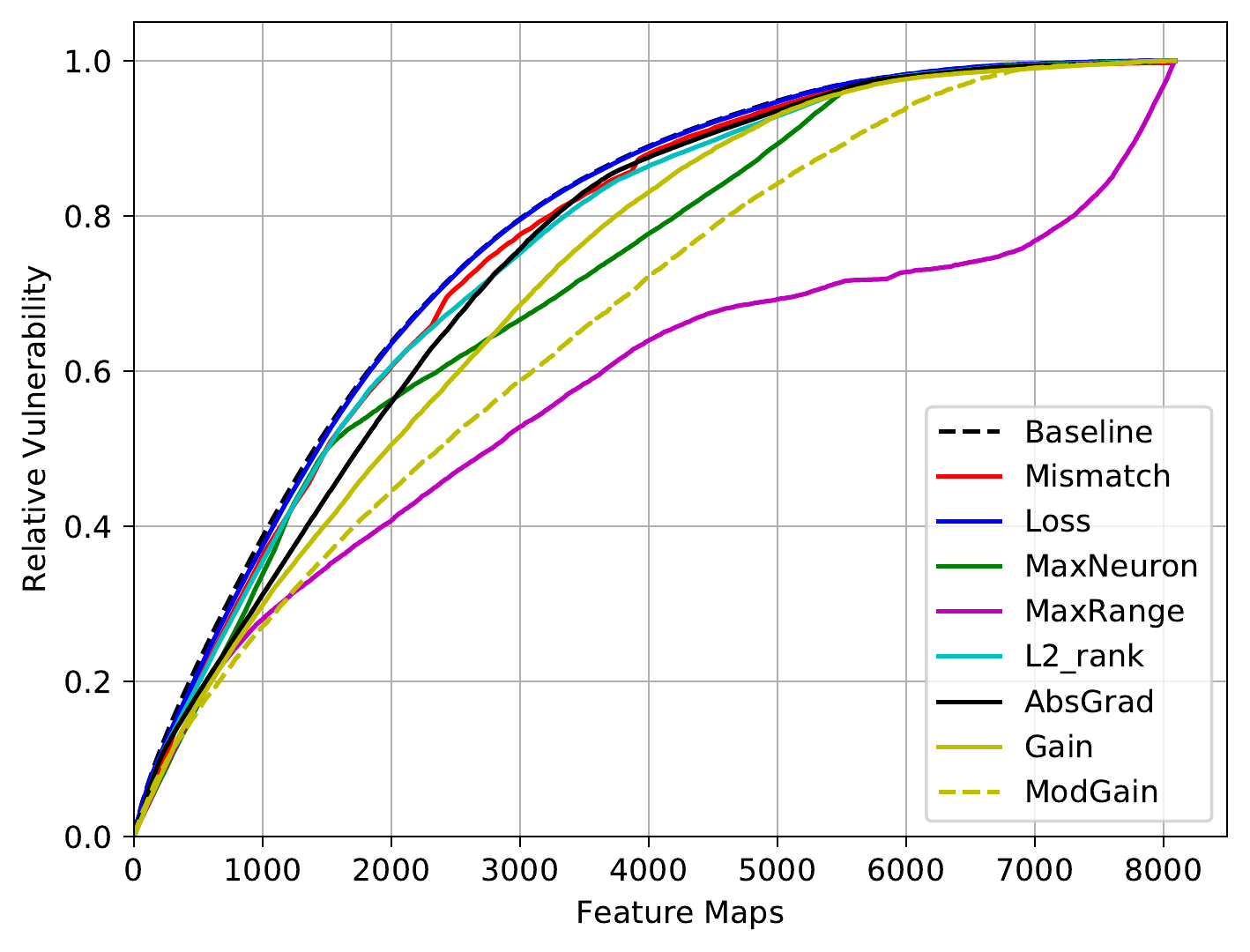}
        \caption{ShuffleNet-ImageNet}
        \label{fig:accuracy:shufflenet}
    \end{subfigure}
        \begin{subfigure}[b] {.33\textwidth}
        \centering
        \includegraphics[width=\textwidth]{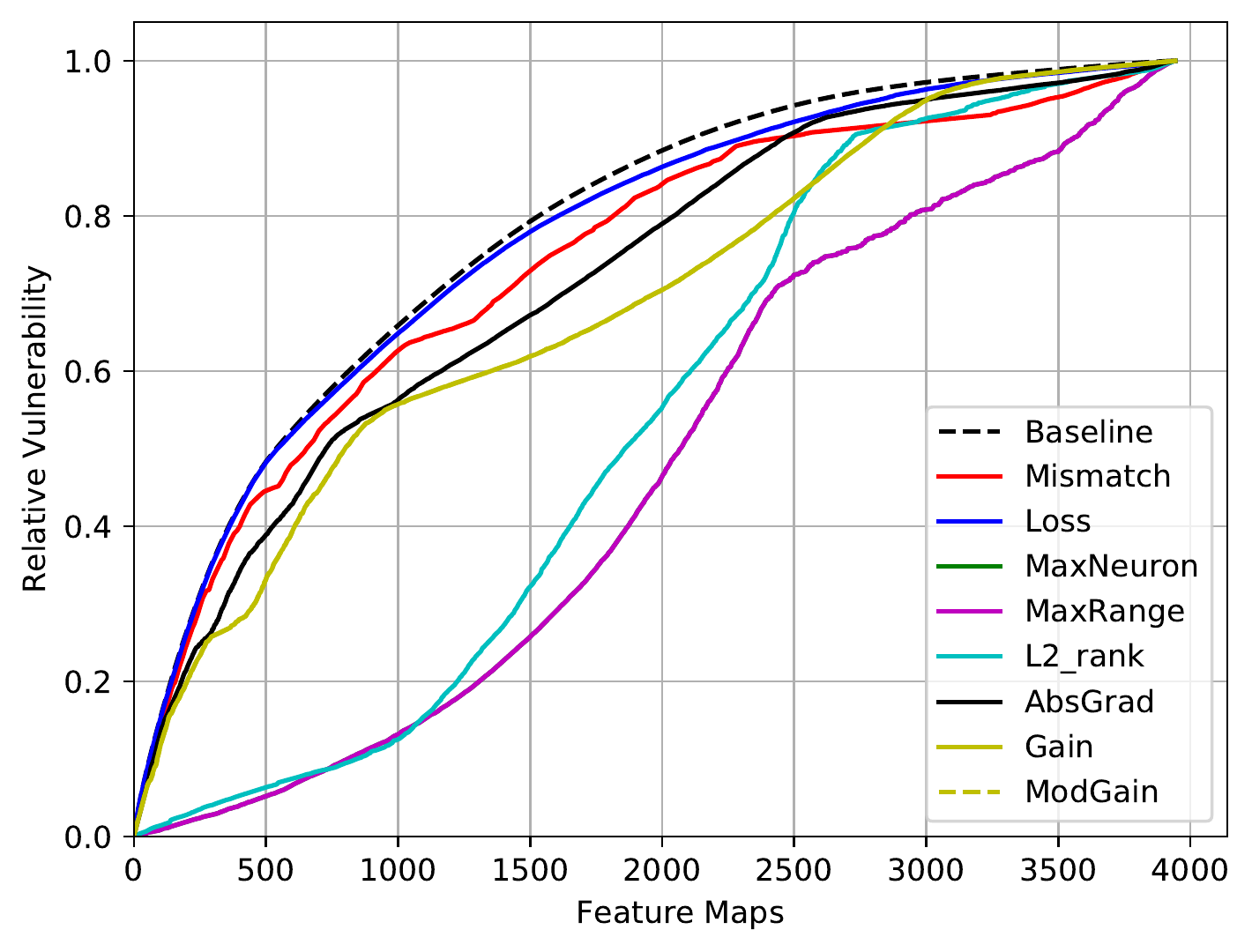}
        \caption{SqueezeNet-ImageNet}
        \label{fig:accuracy:VGG19}
    \end{subfigure}
    \hfill
    \begin{subfigure}[b] {.33\textwidth}
        \centering
        \includegraphics[width=\textwidth]{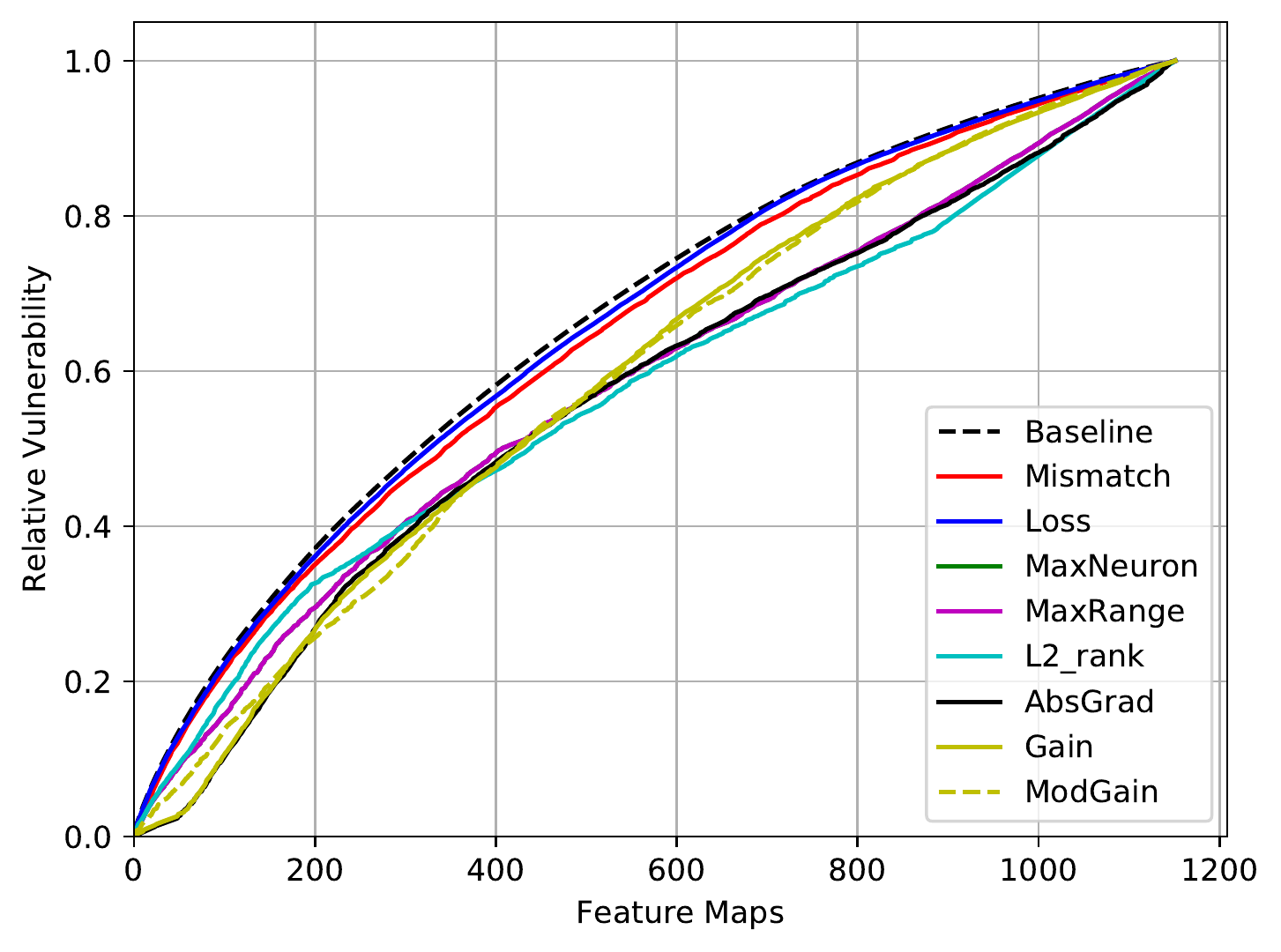}
        \caption{AlexNet-ImageNet}
        \label{fig:accuracy:alexnet}
    \end{subfigure}
    \vspace{-0.2in}
    \caption{Relative vulnerability computed for each heuristic on the ES, compared to baseline relative vulnerability of $\Delta$loss from the TS. Y-axis shows the cumulative sum of of relative vulnerability as obtained by the mismatch on TS. 
    The Manhattan distance between Baseline and a vulnerability estimation technique indicates the accuracy of the method, where shorter distance is better.
    }
    \label{fig:heuristic_accuracy_plots}
    \vspace{-.1in}
\end{figure*}

\subsection{Mismatch vs.\ Loss}
\label{sec:results:mismatch_vs_loss}
We next evaluate the efficacy of using loss-based vulnerability estimation
in place of the mismatch metric to measure relative vulnerability. 
Using AlexNet as a case study, we perform a large injection campaign sweeping the number of injections 
per fmap from 64 to 12288, 
and measure relative fmap vulnerability using both mismatch and $\Delta$loss. We compare
the relative vulnerability at each point to the Oracle vulnerability 
defined as mismatch at 12,288 inj/fmap, and compute the Manhattan distance between the cumulative
relative vulnerabilities. Figure~\ref{fig:mismatch_vs_loss:alexnet} shows the result
for AlexNet across different datasets, and Figure~\ref{fig:mismatch_vs_loss:all_networks} shows
the result for all ImageNet networks with $\Delta$loss compared to Mismatches up to 2048 inj/fmap.

Although the mismatch metric may be considered the gold standard for CNN reliability analysis, we 
find that \textit{$\Delta$loss and mismatch both converge as the total number of injections per fmap increase}.
Larger models, such as AlexNet trained for ImageNet rather than CIFAR, take longer to converge
for mismatch and start much further away from the final result (Manhattan Distance of 0.12). The reason
for this is that the binary nature of the mismatch-based metric (i.e., it must observe a Top-1 misclassification
to differentiate between fmaps) means that more injections are required as a function of the larger space of errors. Thus,
the more total neurons in the CNN, the larger the space of possible errors, which in turn translates to
requiring more error injection experiments if measured by mismatch.

However, $\Delta$loss does not suffer from this phenomenon, as it incorporates small changes
in the output resulting from an error even if the Top-1 class does not change. In other words,
$\Delta$loss can extract information from Masked errors, and use that to quickly converge to
the fmaps final vulnerability estimate. We find this trend across all networks studied for ImageNet
in Figure~\ref{fig:mismatch_vs_loss:all_networks}, where $\Delta$loss asymptotically approaches its
final value quickly (note the log scale on x-axis). For example, for ResNet50, $\Delta$loss using 256 inj/fmap, the relative vulnerability is a distance
of less than .002 from the relative vulnerability of $\Delta$loss using 2048 inj/fmap, indicating a potential runtime improvement of the injection campaign
by 10$\times$ with negligible loss of quality.

As the distance computed in Figure~\ref{fig:mismatch_vs_loss:all_networks}
is with respect to mismatch at 2048, we find that networks with more neurons/fmap 
such as VGG (see column 6 of Table~\ref{table:dnns}) display a larger distance between $\Delta$loss and mismatch.
This can be attributed to the baseline \textit{mismatch} metric not yet converging at 2048 inj/fmap. From Figure~\ref{fig:mismatch_vs_loss:alexnet}, we can expect that as the mismatch metric approaches many more inj/fmap,
the two will close the gap. Thus, $\Delta$loss is in fact a better fine-grained
proxy at estimating fmap vulnerability, since it requires fewer inj/fmaps to converge while still 
maintaining high accuracy compared to mismatch at higher inj/fmaps.

\begin{table}[t]
    \caption{Experimental Runtimes for Vulnerability Estimation (hrs) \\
    *Implementation required batch size = 1 for Gain/Mod-Gain}
	\scriptsize
     	\centering
        \renewcommand{\arraystretch}{1.0}
            \begin{tabular}{||c|l|l|l||}
                    \hline
                      \textbf{Estimation} & \textbf{AlexNet-} & \textbf{GoogleNet-} & \textbf{ResNet50-} \\
                      \textbf{Technique} & \textbf{ImageNet} & \textbf{ImageNet} & \textbf{ImageNet} \\
                    \hline\hline
                    Mismatch-2k     &   1.25     &       8.20      &   35.5  \\
                    \hline
                    $\Delta$Loss-2k     &   1.25     &       8.20      &   35.5  \\
                    \hline
                    MaxNeuron        &   0.01     &       0.01      &   2.50  \\
                    \hline
                    FmapRange        &   0.01     &       0.01      &   2.50  \\
                    \hline
                    Average L2       &   0.01      &      0.01      &   0.02  \\
                    \hline
                    Gradient         &   0.01     &       0.08      &   2.50  \\
                    \hline
                    Gain*            &   0.35     &       3.50      &   15.38  \\
                    \hline
                    Mod-Gain*        &   0.38     &       4.50      &   16.98 \\
                    \hline
        \end{tabular}

\label{table:heuristic_runtime_results}
\end{table}

\subsection{Error Coverage vs.\ Computational Overhead}
\label{sec:results:coverage_vs_overhead}
HarDNN provides the developer with a technique to estimate vulnerability, and tune
for error coverage versus computational overhead. Since the developer may not always
have the true relative vulnerability of each fmap, s/he requires an accurate tool
to make an informed decision regarding the coverage vs.\ overhead tradeoff.
Figure~\ref{fig:predicted_coverage} 
depicts this trade-off, where the x-axis shows the relative vulnerability estimates
using $\Delta$loss, and the corresponding additional percentage of MAC units required
to obtain the corresponding coverage. The coverage here is measured using the PropP value of
$\Delta$loss from the ES for vulnerability estimation, as a prediction of actual coverage
which is provided by the mismatch metric from the TS.

Based on this view, we find that the computational overhead is always
sub-linear to coverage, indicating that selective protection is in fact advantageous 
to full duplication, and can even provide large benefits. For example, covering
90\% of errors in SqueezeNet requires less than 30\% additional MACs. Even for networks
which don't display such a large advantage as SqueezeNet, all networks do exhibit the 
sublinear behavior of coverage to overhead. The reasoning behind this is that the
most vulnerably fmaps have a high PropP and/or a high OriginP. For networks which have less
uniform fmap size distributions, such as SqueezeNet, MobileNet, and ShuffleNet, we find
a knee at the end of the curve which captures large features (OriginP) with low PropP.
Other networks which do not have such a large discrepancy between fmaps across layers
show relative vulnerability with a less pronounced trade-off between coverage and overhead.

We compare the predicted coverage by $\Delta$loss to the actual coverage in 
Figure~\ref{results:validation} (as
described in Section~\ref{sec:eval:coverage_vs_overhead}), and we find that, not only
is $\Delta$loss relatively quick at vulnerability estimation (from Section~\ref{sec:results:mismatch_vs_loss}), 
it is also representative of the actual vulnerability as measured by mismatches in the TS. Thus,
the prediction provided to the developer is very accurate, with $\Delta$loss providing an excellent
proxy for error coverage. This goes back to the fact that $\Delta$loss values actually capture 
the sensitivity of the network, and the mismatch-based metric is a specific by-product of how sensitive
an fmap is to errors.

\subsection{Heuristic Analysis}
\label{sec:results:heuristic_analysis}
Last but not least, we evaluate the accuracy and runtime of all heuristics, 
measured against the vulnerability estimate of $\Delta$loss 
as a baseline since $\Delta$loss experimentally proves to be a 
superior metric to mismatches. Figure~\ref{fig:heuristic_accuracy_plots} shows the heuristic
results for different CNNs, where the x-axis is ordered based on the relative vulnerability
estimate for each respective heuristic using the ES, and the baseline for $\Delta$loss is provided
by the TS. 
The relative vulnerability is obtained for
each heuristic by extracting the PropP of each Fmap from the mismatch-based TS, and the cumulative
sum is shown on the y-axis. Table~\ref{table:heuristic_runtime_results} 
shows the measured runtimes for all the techniques on our evaluation infrastructure.  

We find that $\Delta$Loss is the highest performing metric (average distance of .004 from baseline)
followed by mismatch (average distance = .006), both from the ES. This validates that the error injection 
techniques have high accuracy relative to the TS. 
The non-injection based techniques however vary widely, but we find that the techniques which leverage backprop are generally more accurate, with \textit{Gradient} performing the best overall (average distance = .067). This result follows from the
fact that the magnitude of the gradient helps inform the overall sensitivity of the fmap to a perturbation; a gradient
of zero means the fmap is very stable, and won't change too much, translating to a smaller effect on the output. Larger
gradients on the other hand may play a stronger role in the final classification for an image, and as such the backprop-based
metrics leverage this information. 

One additional insight we find is that on average, a small number of fmaps (less than a third) account for a large
percentage of the networks relative vulnerability (average of 76\% cumulative vulnerability). 
However, this does not directly map to the overhead associated with the highly vulnerable feature maps (as it doesn't take into
account the size of the fmap, just the number of fmaps) but indicates that without incorporating OrigP, the
relative vulnerability of fmaps may be biased.

For runtime analysis of the heuristics, we leveraged image batching when available, and still 
found that the runtime trends between the analytical models (Section~\ref{sec:eval:heuristic_analysis}) 
and the measure runtimes to be similar. However, gain and mod-gain underperformed due to
an implementation limitation, where the backprop algorithm did not support for batching with different differentiation
values as required by the gain formulation. 
Overall, $\Delta$loss provides the highest accuracy and be sped up with fewer inj/fmap, while
gradient provides an acceptable trade-off between runtime and accuracy, 
which can be used for quick profiling by the developer during vulnerability estimation.

\section{Related Work}

\textbf{Pruning Based Techniques}:
CNN model pruning techniques aim to remove redundant and less-useful parameters from
a model to improve execution efficiency~\cite{CunBrainDamage}. These techniques often reduce accuracy by a few small factors. 
HarDNN focuses on identifying
vulnerable feature maps, which it then proceeds to duplicate to improve reliability, with no effect on classification accuracy.
There are many similarities between pruning and hardening. (1) Pruning is typically a two-phased process. The first phase identifies a filter to remove, and the second phase (called a fine-tuning phase) removes the filter and retrains the network. (2) Recent work found that pruning full filters 
(rather than individual weights in a filter) can have minimal effects on accuracy, while improving
the pruning speed~\cite{pruning_1}. This is analogous to our 
fmap target granularity and protection strategy. 
(3) Pruning techniques rank filters using heuristics to identify candidates to prune~\cite{pruning_1, pavloPruning}. 
We also explore similar heuristics to estimate fmaps based on vulnerability. 
The objective of a pruning technique is to zero-out a filter,
removing it from the model. In contrast, for the resiliency analysis, we assume various error models which change a single neuron.





\textbf{DNN Reliability}: Recent work has explored DNN-specific reliability due to rise of DNN usage 
in safetly-critical applications. Previous methods targeted neuron-level~\cite{SchornGuntoro2018} 
vulnerability but more recently have also gravitated toward fmap-level analysis~\cite{SchornGuntoroDATE19}. 
HarDNN differs from Schorn et al. in that their focus is on redistributing
error across a CNN, whereas HarDNN aims to provide selective protection of the vulnerable fmaps. BinFI~\cite{BinFI} proposes
an orthogonal binary search technique to reduce the error injection space for ML reliability, which can be generally
used to speed up error injection campaigns. We introduce $\Delta$loss as a different metric for measuring vulnerability,
which, as shown, can also speed up injection campaigns.

\section{Conclusion}
This paper presents HarDNN, a software-directed technique to identify vulnerable 
computations in CNNs and selectively protect them. HarDNN operates at the feature
map level granularity, and introduces $\Delta$loss as an accurate 
error-injection based metric for vulnerability estimation, and explores different heuristics
for fast vulnerability assessment. Additionally, we analyze the tradeoff between error
coverage and computation overhead for selective protection. Results show that 
the relative vulnerability of an fmap is similar across 3 error models studied,
and that the improvement in resilience for the added computation is super linear with HarDNN.
For example, HarDNN can improve SqueezeNet's resilience by 10$\times$ with just 30\% computational overhead. 
For future work
we plan to extend HarDNN to include other applications of neural networks.

\section*{Acknowledgements}
This material is based upon work supported in part by the Applications Driving Architectures (ADA) Research Center, a JUMP Center co-sponsored by SRC and DARPA. A portion of this
work was performed while Abdulrahman Mahmoud interned at NVIDIA.

\bibliography{bibs/sadve1,bibs/sadve2,bibs/sadve3,bibs/sadve4,bibs/hardnn}
\bibliographystyle{sysml2019}

\end{document}